\documentclass{article} 
\usepackage{iclr2023_conference,times}
\usepackage{amssymb}
\usepackage{tabularx}
\usepackage{amsthm}
\usepackage[ruled,linesnumbered]{algorithm2e}
\usepackage{mathtools}
\usepackage{xcolor, soul}
\usepackage{subfigure}

\definecolor{table_color}{rgb}{0.769, 0.922, 0.984}
\usepackage{graphicx}

\usepackage{amsmath,amsfonts,bm}

\def\eqref#1{equation~\ref{#1}}

\def\1{\bm{1}}

\DeclareMathAlphabet{\mathsfit}{\encodingdefault}{\sfdefault}{m}{sl}
\SetMathAlphabet{\mathsfit}{bold}{\encodingdefault}{\sfdefault}{bx}{n}

\newcommand{\E}{\mathbb{E}}

\usepackage{hyperref}
\usepackage{url}

\theoremstyle{plain} 
\theoremstyle{plain} \newtheorem{Theorem}{Theorem}

\title{Offline Reinforcement Learning \\ with Adaptive Behavior Regularization}

\author{Yunfan Zhou \\
Equal contributions\\
Chinese University of Hong Kong (Shenzhen)\\
Shenzhen, China \\
\AND
Xijun Li\thanks{Xijun Li is the corresponding author.} \\ 
Equal contributions\\
University of Science and Technology of China \\
Hefei, China \\
\AND
Qingyu Qu \\
Beihang University \\
Beijing, China
}

\iclrfinalcopy 
\begin{document}

\maketitle

\begin{abstract}
Offline reinforcement learning (RL) defines a sample-efficient learning paradigm, where a policy is learned from static and previously collected datasets without additional interaction with the environment. The major obstacle to offline RL is the estimation error arising from evaluating the value of out-of-distribution actions. To tackle this problem, most existing offline RL methods attempt to acquire a policy both ``close" to the behaviors contained in the dataset and sufficiently improved over them, which requires a trade-off between two possibly conflicting targets. In this paper, we propose a novel approach, which we refer to as \textit{adaptive behavior regularization (ABR)}, to balance this critical trade-off.
By simply utilizing a sample-based regularization, ABR enables the policy to adaptively adjust its optimization objective between cloning and improving over the policy used to generate the dataset.
In the evaluation on D4RL datasets, a widely adopted benchmark for offline reinforcement learning, ABR can achieve improved or competitive performance compared to existing state-of-the-art algorithms. 


\end{abstract}

\section{Introduction}
Reinforcement learning (RL) has achieved remarkable advances over various domains in recent years~\citep{sutton2018reinforcement, mnih2013playing, silver2017mastering, kalashnikov2018scalable}. Traditional reinforcement learning requires active interactions with the environment to explore the state-action space. However, interacting with the environment using a stochastic and possibly suboptimal policy can be costly and dangerous, especially in real-world scenarios ~\citep{tejedor2020reinforcement, yu2021reinforcement, kiran2021deep}. As a compelling alternative, offline RL ~\citep{fujimoto2019off, levine2020offline} was proposed to eliminate online interactions with the environment. In offline RL, the policy is learned from a fixed dataset previously collected by arbitrary and usually unrevealed behavior policies. Without additional online data collecting, the offline RL is thus able to learn policies more efficiently and safely.



However, the absence of online interaction also makes offline RL a challenging task. Though most value-based off-policy RL algorithms are applicable to offline tasks, they usually perform poorly in practice.
The major obstacle is the estimation error caused by \textit{distributional shift}~\citep{levine2020offline}, which is essentially the discrepancy between the learned policy and the behavior policy used to generate the datasets.
Specifically, when a learned policy is improved over the behavior policy, it tends to come up with actions distinct from the behaviors observed in the dataset, resulting in a distributional shift.
Accurately evaluating the value of actions not contained in the datasets can be tricky since it is hard to rectify the incorrect estimation without the feedback of online interaction. The erroneous estimation then misleads the learned policy to prefer actions with over-estimated value, harming the performance in offline tasks.  


To mitigate the impact of the erroneous estimation caused by distributional shift, a straightforward method is to restrict the learned policy ``close'' to the behavior policy. Previous works generally achieve this goal by imposing a constraint on the learned policy~\citep{fujimoto2019off, kumar2019stabilizing, peng2019advantage, wu2019behavior, fujimoto2021minimalist}, regularizing the estimated value functions~\citep{kumar2020conservative, kostrikov2021offlinefisher}, or adding penalties to actions with highly uncertain value\citep{bai2022pessimistic, wu2021uncertainty, an2021uncertainty}. 
However, while restricting the learned policy effectively avoids querying the value of out-of-distribution actions, it possibly forces the learned policy to deviate from the optimal one~\citep{kostrikov2021offline, levine2020offline, kostrikov2021offlinefisher}. Hence, for these offline RL methods, there is a critical trade-off between restricting the learned policy to stay ``close" with the behavior policy and improving the learned policy to obtain the optimal expected value.

To balance this crucial trade-off, we propose a novel approach, namely \textit{adaptive behavior regularization (ABR)}, that enforces a constraint with adaptive weight to the learned policy. As in the case of many previous works~\citep{wu2019behavior, kumar2019stabilizing, fujimoto2021minimalist}, we construct a policy improvement objective with a constraint on the discrepancy between the learned policy and the behavior policy. Nevertheless, it significantly differs from previous methods since the weight of the constraint is adaptively adjusted, depending on how much the learned policy deviates from the behavior policy. Hence, the learned policy is improved following different patterns during the training process. When it deviates far from the behavior policy, it tends to clone the behavior policy. When sufficiently ``close" to the behavior policy, it is approximately improved towards maximizing the expected return. ABR can be simply implemented based on a standard actor-critic framework and is easy to tune. In the evaluation on D4RL~\citep{fu2020d4rl}, a 
popular benchmark for offline RL, our approach outperforms many previous state-of-the-art offline algorithms in most tasks. Particularly, our approach significantly improves over the previous algorithms on datasets with complex and multimodal distribution.

\section{Preliminaries}
\label{preliminaries}
\textbf{Reinforcement Learning}. Reinforcement learning ~\citep{sutton2018reinforcement} is defined in the context of Markov Decision Process (MDP) represented by a tuple $(\mathcal{S}, \mathcal{A}, T, r, \gamma)$, where $\mathcal{S}$, $\mathcal{A}$ represent space of states and actions respectively, $T(s'\vert s, a)$ represents environment dynamics, $r(s, a)$ represents reward function, and $\gamma$ represents discount factor. An Agent decides its action by a policy function, $\pi(a\vert s)$, mapping from states to actions. State visitation frequency induced by $\pi$ is represented by $d^{\pi}(s)$. The goal of RL is to obtain the optimal policy that maximizes the expectation of return, namely cumulative discounted rewards, which is measured by Q-function. The Q-function following policy $\pi(a\vert s)$ is defined as: 
\begin{equation}
    \begin{aligned}
    Q^{\pi}(s, a) \triangleq \mathbb{E}_{\substack{s_{t+1} \sim T(\cdot\vert s_t, a_t)\\ a_t \sim \pi(\cdot \vert s)}} \left[\sum \limits_{t=0}^{\infty} \gamma^t r(s_t, a_t)\right]
    \end{aligned}
\end{equation}

\textbf{Offline RL}. In offline RL, datasets are static and previously-collected. An offline Dataset is defined as a set of transitions: $\mathcal{D}=\{(s, a, s', r)\}$. We assume that offline datasets are sampled by an unknown behavior policy $\pi_{\beta}$.
Offline RL based on approximate dynamic programming are typically based on actor-critic framework, which alternates between policy evaluation and policy improvement. In the policy evaluation phase, $Q^{\pi}$ is calculated via Bellman backup~\citep{sutton2018reinforcement}:  $\mathcal{B}^{\pi} Q(s,a) = r(s,a) + \gamma\mathbb{E}_{s' \sim T(\cdot\vert s, a)a' \sim \pi(\cdot \vert s)}\left[Q(s', a')\right]$, where $\mathcal{B}^{\pi}$ is the Bellman operator following $\pi$. 
The objective for policy improvement is to maximize expected Q-value:
\begin{equation}
\label{policy improvement}
\begin{aligned}
    & \text{Policy evaluation:} \quad
    \hat{Q}^{k+1} \leftarrow \mathop{\arg \min}_{Q}  \left[Q(s, a) - \mathcal{B}^{\hat{\pi}^k}\hat{Q}^k(s, a)\right]\\
    & \text{Policy improvement:} \quad 
    \hat{\pi}^{k+1} \leftarrow \mathop{\arg \max}_{\pi} \mathbb{E}_{\substack{s \sim \mathcal{D} \\a\sim \pi(\cdot \vert s)}} \left[\hat{Q}^{k+1}(s, a)\right]
\end{aligned}
\end{equation}

One central challenge for offline RL is the erroneous estimation caused by distributional shift, which is essentially the inability of function approximators to evaluate a policy that differs substantially from $\pi_{\beta}$. 
For the algorithms based on standard actor-critic framework, while the Q-function approximator is only trained on the actions sampled from the datasets, it has to deduce the Q-value of actions sampled by $\hat{\pi}^k$ in both policy evaluation and policy improvement phase. Since the learned policy is improved towards maximizing the expected Q-value, it may prefer actions with over-estimated Q-value, resulting in poor performance. 


\section{Related Works}
\label{related work}
In this section, we mainly introduce prior works in model-free offline RL.
A substantial portion of existing offline RL methods address the distributional shift problem by constraining the deviation from the behavior policy to the learned policy, ensuring the out-of-distribution actions are rarely queried to the Q-function. 
This can be typically achieved by imposing a constraint on the policy improvement objective, where the deviation from the behavior policy to the learned policy can be measured in terms of KL-divergence~\citep{wu2019behavior, jaques2019way, peng2019advantage, siegel2020keep}, MMD distance~\citep{kumar2019stabilizing}, Wasserstein distance~\citep{wu2019behavior}, etc. The constraint can be accomplished explicitly by adding a penalty term to the policy improvement objective~\citep{kumar2019stabilizing, wu2019behavior, fujimoto2021minimalist}, or implicitly via a weighted maximum likehood~\citep{nair2020accelerating, wang2020critic, peng2019advantage}. We will refer to this category of methods as policy constraint methods~\citep{levine2020offline} in this paper. 

Aside from imposing a constraint on the policy improvement objective, an effective alternative is regularizing the Q-function to avoid overestimating the value of out-of-distribution actions~\citep{kumar2020conservative, kostrikov2021offline}. These methods usually aim to obtain a Q-function that produces relatively low Q-value for out-of-distribution actions. Since the policy is improved to maximize the regularized Q-function, the learned policy is then trained to prefer in-distribution actions. Besides, some algorithms have proposed to mitigate distributional shift by leveraging uncertainty estimation. These methods attempt to obtain a policy that only covers the trustworthy action space where the value is with relatively low uncertainty. Quantification of uncertainty has been well studied in online reinforcement learning~\citep{auer2008near, osband2016deep, burda2018exploration}, and introduced to offline reinforcement learning ~\citep{bai2021principled, bai2022pessimistic, wu2021uncertainty, an2021uncertainty}.

Our proposed method, which we refer to as \textit{Adaptive Behavior Regularization (AWR)}, is mostly related to the policy constraint methods. We also obtained a constrained policy improvement objective as these methods. However, our method significantly differs from previous policy constraint methods due to the adaptive regularization weight, which is determined by the probability density of $\pi_{\beta}$. In our method, the weight of the constraint increases when the deviation from the behavior policy to the learned policy does. Besides, in contrast to most policy constraint methods that add a regularizer to the policy improvement objective, we obtain the constrained policy improvement objective by regularizing the policy evaluation.


\section{adaptive behavior regularization}

\subsection{Trade-off in Offline RL}
\label{trade-off}

A critical trade-off in offline RL lies between two contradictory targets. 
On the one hand, the whole point of offline RL is to learn a policy distinct from, and possibly improved over, the behavior policy that generates the datasets.
On the other hand, the learned policy is required to stay “close” with the behavior policy to prevent out-of-distribution action queries to Q-function~\citep{levine2020offline}. 
To further illustrate this trade-off, we study a general case in offline RL, where the policy is trained to maximize the Q-function with a soft constraint on the deviation between the behavior policy and the learned policy. We denote a parameterized policy with parameters $\theta$ as $\pi_{\theta}$ and a parameterized Q-function approximator with parameters $\phi$ as $\hat{Q}_{\phi}$.
In each gradient step, the parameters $\theta$ are updated towards maximizing:
\begin{equation}
\label{soft constraint}
    \begin{aligned}
        J(\theta) \triangleq \mathbb{E}_{\substack{s \sim \mathcal{D} \\a\sim \hat{\pi}_{\theta}(\cdot \vert s)}}
        \left [\hat{Q}_{\phi}(s, a) - \alpha f(s, a)\right]
   \end{aligned},
\end{equation}
where $\mathbb{E}_{s \sim \mathcal{D}, a\sim \hat{\pi}_{\theta}(\cdot|s)}f(s,a)$ is a \textit{behavior regularizer} that measures the deviation from the behavior policy to the learned policy, i.e., the soft constraint term. This soft constrained objective function is widely adopted in previous works~\citep{wu2019behavior, fujimoto2021minimalist, peng2019advantage}, where the choice of $\alpha$ explicitly represents the trade-off. 
To estimate the gradient of $J(\theta)$, we transform the expression of a stochastic policy using the reparameterization trick~\citep{zhang2018advances}:
$
    a = g_{\theta}(\epsilon; s)
$,
where $\epsilon$ is a random variable sampled from a fixed distribution. We now can represent the gradient of $J(\theta)$ based on the chain-rule gradient computation as follows:
\begin{equation}
\label{gradient}
    \nabla_{\theta} J(\theta) =  \nabla_{a}\hat{Q}_{\phi}(s,a)\nabla_{\theta}g_{\theta}(\epsilon; s) -\alpha\nabla_{a} f(s, a)\nabla_{\theta}g_{\theta}(\epsilon; s),
\end{equation}
where the first term is towards cloning the behavior policy and the second is towards maximizing the expected return. When an appropriate weight $\alpha$ is chosen, optimizing the objective in Equation~(\ref{soft constraint}) offers an attractive option to reconcile those two contradictory objectives. 

However, the policy obtained by optimizing Equation~(\ref{soft constraint}) still suffers from obstacles caused by the trade-off. The major limitation is that it is prone to be \textit{over-regularized} (see section~\ref{Discussion} for empirically demonstration). We can take a deeper look at how the policy is updated using sampled actions by rewriting Equation~(\ref{gradient}) in a piece-wise form:
\begin{equation}
\label{piece-weise gradient}
\nabla_{\theta} J(\theta) = 
    \begin{cases}
         \nabla_{a}\hat{Q}_{\phi}(s,a)\nabla_{\theta}g_{\theta}(\epsilon; s) - \underbrace{ \alpha\nabla_{a} f(s, a)\nabla_{\theta}g_{\theta}(\epsilon; s)}_{bias},  
        &\pi_{\beta}(a|s)>0 ;\\
         \underbrace{ \nabla_{a}\hat{Q}_{\phi}(s,a)\nabla_{\theta}g_{\theta}(\epsilon;s)}_{uncertainty}-\alpha\nabla_{a} f(s, a)\nabla_{\theta}g_{\theta}(\epsilon; s), 
        &\pi_{\beta}(a|s)=0.
    \end{cases}
\end{equation}
As shown in Equation~(\ref{piece-weise gradient}), if an out-of-distribution action is sampled, the evaluated Q-value is highly uncertain, possibly misleading the learned policy to prefer actions with over-estimated Q-value. To tackle this problem, the weight of the behavior regularizer is required to be sufficiently large to dominate the objective, such that the policy is instead updated towards minimizing the deviation from the behavior policy to the learned policy. However, in the other case that an in-distribution action is sampled, a large $\alpha$ means a considerable bias is introduced to its Q-value. This will force the learned policy to update towards the behavior policy instead of the optimal one, resulting in over-regularization on the learned policy.

To alleviate this limitation, we can, in principle, construct a policy improvement objective with an adaptive balance between the Q-function and the behavior regularizer, whose gradient function takes the form of:
\begin{equation}
\label{adaptive weight}
    \begin{aligned}
        \nabla_{\theta} \tilde{J}(\theta) \triangleq 
       ( 1-\alpha(s,a)) \nabla_{a}\hat{Q}_{\phi}(s,a)\nabla_{\theta}g_{\theta}(\epsilon; s) - 
       \alpha(s,a) \nabla_{a}f_{\theta}(a)\nabla_{\theta}g_{\theta}(\epsilon; s)
   \end{aligned},
\end{equation}
where $\alpha(s, a)$ is ranged in $[0, 1]$, adjusted depending on the discrepancy between the behavior policy and the learned policy. Specifically, for actions significantly distinct from the behaviors contained in the datasets, the $\alpha(s, a)$ is large. Thus, the weight of the regularizer exceeds that of the Q-function. When these actions are sampled, the policy tends to update towards minimizing the deviation to the behavior policy, limiting the effect caused by erroneous estimation to Q-function. On the contrary, the $\alpha(s, a)$ is sufficiently small for actions similar to the behavior contained in the datasets. Then, the gradient of the policy improvement objective approximates that of the Q-function, thus effectively mitigating the over-regularization. 

\subsection{Adaptively Regularized Policy Improvement}
\label{sec: ABR framework}
In this section, we propose a simple yet effective offline RL method, which we refer to as \textit{adaptive behavior regularization} (ABR). All proofs for the derivation in this section can be found in Appendix~\ref{theoretical proofs}. 
In ABR, we impose an adaptive constraint, which restricts the discrepancy between the learned policy and the behavior policy with an adaptive weight to the policy improvement.
First, we present a simple regularization method to produce an adaptive weight by assigning a surrogate value $\tilde{Q}(s,a)$ to the actions uniformly sampled from the action space. 
A family of policy evaluation objective characterized by the $\tilde{Q}(s,a)$ then can be defined as Equation~(\ref{Q-function iteration}):
\begin{equation}
\label{Q-function iteration}
    \hat{Q}_{\phi}^{k+1}\leftarrow  \mathop{\arg \min}_{Q}
    \mathbb{E}_{s\sim d^{\pi_{\beta}}, a\sim \pi_{\beta}}
    [Q(s,a)-\mathcal{B}^{\hat{\pi}^k}\hat{Q}^{k}(s,a)]^2 + 
    \alpha\mathbb{E}_{s\sim d^{\pi_{\beta}},a\sim \mathcal{U}}[Q(s,a)-\tilde{Q}(s,a)]^2,
\end{equation}
where $\mathcal{U}$ is a uniform distribution over the action space, and $\alpha$ is the coefficient
of the regularizer. 

\textbf{Q-function iteration.} We denote the Q-function iteration in Equation~(\ref{Q-function iteration}) as $Q(s, a) \leftarrow\tilde{\mathcal{B}}^{\pi}Q(s,a)$ and the probability density of $\mathcal{U}$ by a constant $u$. 
Under the assumption that, in the $k$th training iteration, we obtain the learned policy by greedily solving the optimization problem in Equation~(\ref{Q-function iteration}), we can get $\hat{Q}^{k+1}_{\phi}$ in terms of $\hat{Q}^{k}_{\phi}$:
\begin{equation}
\label{Q-value}
\begin{aligned}
\forall s \in \mathcal{D},a \in \mathcal{A} \quad
\hat{Q}_{\phi}^{k+1}(s,a)=\left(1 - \frac{\alpha u}{\pi_{\beta}(a|s)+\alpha u}\right) \mathcal{B}^{\hat{\pi}^k}\hat{Q}_{\phi}^{k}(s,a) + \frac{\alpha u}{\pi_{\beta}(a|s)+\alpha u}\tilde{Q}(s,a).
\end{aligned}
\end{equation}
We thus construct an adaptive weight that monotonically decreases as the $\pi_{\beta}(a|s)$ increases, as in the case we discussed in Section~\ref{trade-off}. Moreover, this adaptive weight is independent on an explicit estimation to $\pi_{\beta}(a|s)$. We highlight that the absence of estimating $\pi_{\beta}(a|s)$ is a significant advantage of ABR
since modeling the behavior policy is usually challenging, especially when the datasets comprise experience from multiple policies. In our experimental evaluation (see section~\ref{offline evaluation}), we observe that ABR outperforms the existing offline RL methods in most datasets generated by complex behavior policies.

To compare the policy improvement objective with adaptive regularization weight to that with a fixed weight, we rewrite Equation~(\ref{Q-value}) in a piece-wise form and study the regularized Q-function of in-distribution and out-of-distribution actions respectively:
\begin{equation}
\label{piece-wise Q-function}
\hat{Q}_{\phi}^{k+1}(s,a)= 
\begin{cases}
 \mathcal{B}^{\hat{\pi}^k}\hat{Q}_{\phi}^{k}(s,a) - \frac{\alpha u}{\pi_{\beta}(a|s)+\alpha u}(\mathcal{B}^{\pi}\hat{Q}_{\phi}^{k}(s,a) - \tilde{Q}(s,a)), &\quad \pi_{\beta}(a|s) > 0;\\
 \tilde{Q}(s,a), &\quad \pi_{\beta}(a|s)=0.
\end{cases}
\end{equation}

In Equation~(\ref{piece-wise Q-function}), we present the regularized Q-function in and out of the support of $\pi_{\beta}(a|s)$. In the case that $\pi_{\beta}(a|s)>0$, the regularized Q-function consists of the Q-function calculated by standard Bellman backup and a bias term. While $\pi_{\beta}(a|s)=0$, the regularized Q-value is equal to $\tilde{Q}(s,a)$. We then study both cases in detail to complete the design of ABR.

\textbf{Regularized Q-value of out-of-distribution actions.} As shown in Equation~(\ref{piece-wise Q-function}), for any $(s, a)$ pair not contained in the dataset, the regularized Q-value is simply represented by $\tilde{Q}(s, a)$. Since the gradient of the regularized Q-function is supposed to approximate that of the behavior regularizer, $\tilde{Q}(s,a)$ can be naturally defined as:
\begin{equation}
    \tilde{Q}(s, a) =  c(s) - f(s, a),
\end{equation}
where $c(s)$ is a term independent on $a$, to reduce the variance in Q-function estimation (we discuss this in Appendix~\ref{variance}). 

When an out-of-distribution action is sampled, updating the policy towards maximizing its regularized Q-value is equivalent to minimizing the deviation between the learned policy and the behavior policy. Since the regularized Q-value of out-of-distribution actions is determined by $\tilde{Q}(s,a)$, the impact of the erroneous estimation on policy improvement is effectively alleviated.
In our implementation, we define $c(s)$ as the expectation of the Q-value in state $s$ following $\pi_{\beta}(\cdot|s)$, so as to reduce the variance in Q-function estimation. Accordingly, in the $k$th iteration,  the $\tilde{Q}$ is constructed as:
\begin{equation}
    \tilde{Q}(s, a) =  \mathbb{E}_{a \sim \pi_{\beta}(\cdot|s)}\left[\mathcal{B}^{\pi}\hat{Q}^k(s,a)|s\right] - f(s, a),
\end{equation}

\textbf{Regularized Q-value of in-distribution actions.} In Theorem~\ref{bounded bias}, we show that the bias introduced to the evaluated Q-value is bounded. Specifically, if we give a bounded $\tilde{Q}(s,a)$, then for any in-distribution actions where $\pi_{\beta}(a|s)$ is sufficiently large, the bias is bounded by a term proportional to $\alpha$. This suggests that we can always reduce the bias by setting a sufficiently small $\alpha$. Meanwhile, reducing the scale of $\alpha$ in principle does not affect the regularized Q-value of out-of-distribution actions. Hence, when adopting an appropriate $\alpha$, the regularized Q-function approximates the standard Q-function in the in-distribution actions, while approximating the behavior regularizer in the out-of-distribution actions.
Proofs of Theorem~\ref{bounded bias} can be found in Appendix~\ref{theoretical proofs}.
\begin{Theorem}
\label{bounded bias}
(Bounded bias estimation) For any $(s,a)$ that satisfies $\pi_{\beta}(a|s)>\sigma$, given that the reward $r$ is bounded by $r \in [-R_{max}, R_{max}]$, $\tilde{Q}(s,a)$ is bounded by $\tilde{Q}(s,a) < \delta$, for any $s$, $a$, then:
\begin{equation}
    |\tilde{\mathcal{B}}^{\pi}Q(s,a) - \mathcal{B}^{\pi}Q(s,a)| < \frac{\alpha u}{\sigma}(\frac{R_{max}}{1-\gamma} + \delta)
\end{equation}
\end{Theorem}

In practice, we can take different $f(s,a)$ that produces bounded $\tilde{Q}(s, a)$ as a measurement of the distance between $\pi_{\beta}(\cdot|s)$ and $\pi_{\theta}(\cdot|s)$. We prefer to simply define $f(s,a) = \mathbb{E}_{a'\sim \pi_{\beta}(\cdot|s)}\left[\lambda(a - a')^2|s\right]$, where $\lambda$ is a constant coefficient, to avoid directly estimating $\pi_{\beta}(\cdot|s)$. When we take this definition and use a stochastic policy with a fixed noise, maximizing $\tilde{Q}(s,a)$ monotonically reduces the generalized energy distance~\citep{rizzo2016energy} between the behavior policy and the learned policy. 
 
\subsection{Discussion}
\label{Discussion}
In this section, we present some intuitive understandings behind ABR and demonstrate it empirically. ABR can be regarded as a novel alternative to prior methods that enforces a constraint on policy. Nevertheless, it significantly differs from prior methods due to \textit{adaptive policy constraint weight}. Specifically, in Equation~(\ref{Q-function iteration}), the evaluated Q-value can be regarded as a weighted average of the Q-value calculated by standard Bellman backup, and the policy constraint term $\tilde{Q}(s,a)$. With any given $\alpha$, the behavior policy determines the weight of the policy constraint term, converging from 1 ($\pi_{\beta}(a|s)=0$) to 0 ($\pi_{\beta}(a|s) \rightarrow +\infty$). In other words, this objective constrains the policy with stronger strength when more out-of-distribution actions are sampled. On the contrary, when in-distribution actions are sampled, the objective enables the policy to maximize the expected return with a mild constraint. 
\begin{figure}[t]
\centering
    \subfigure[TD3+BC]{
        \begin{minipage}[b]{0.4\textwidth}
        \includegraphics[width=1.0\columnwidth]{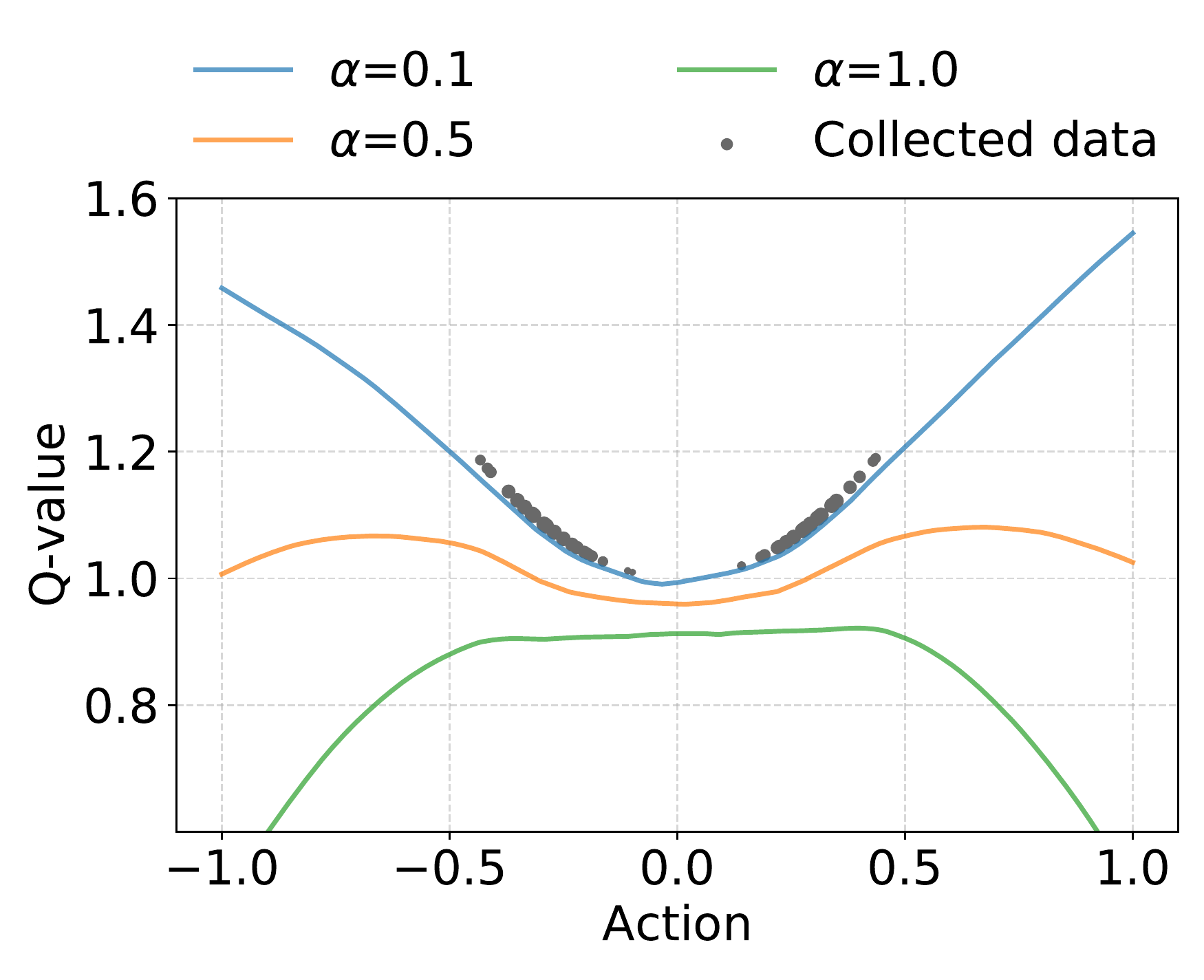}
        \end{minipage}
    }
    \qquad
    \subfigure[ABR]{
        \begin{minipage}[b]{0.4\textwidth}
        \includegraphics[width=1.0\columnwidth]{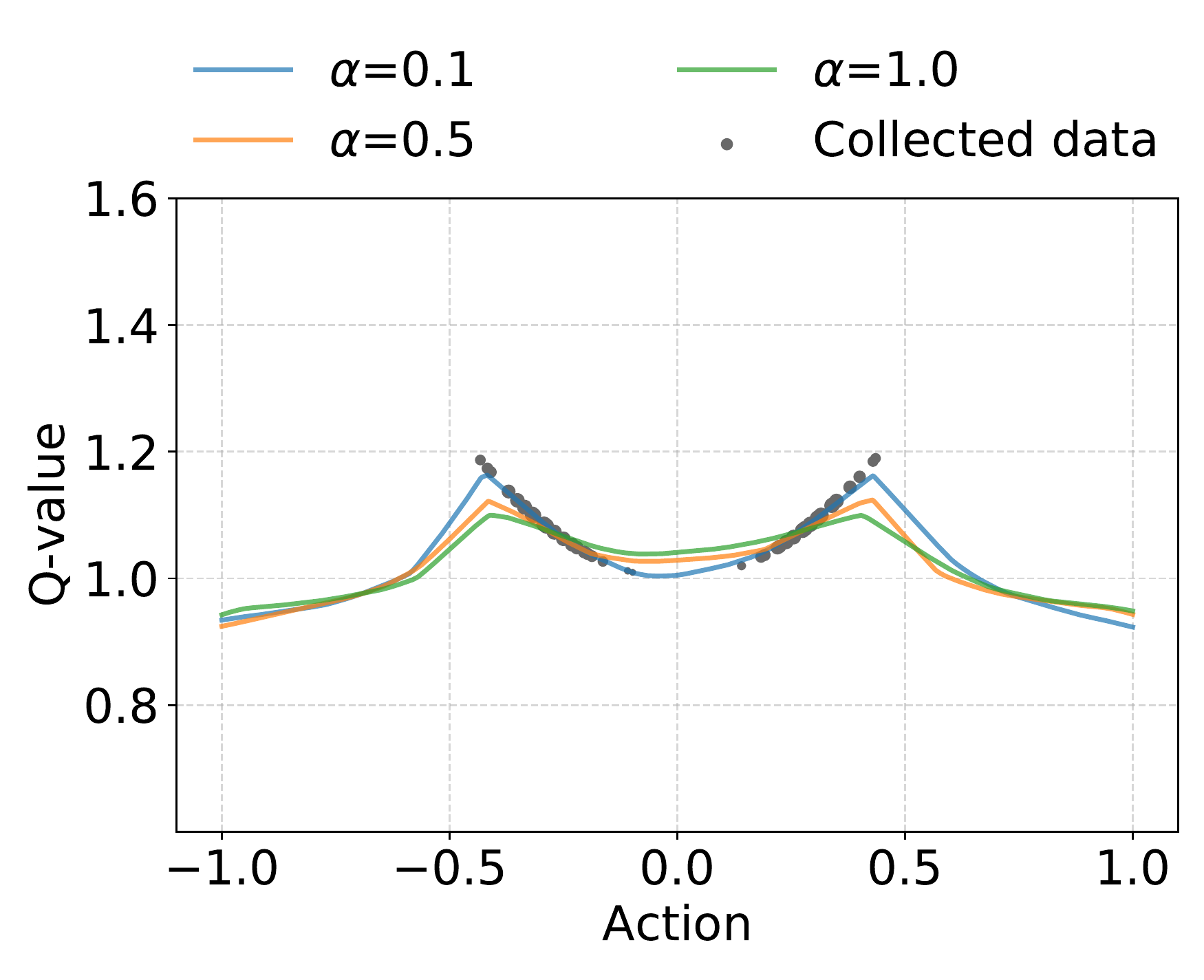} 
        \end{minipage}
    }
\caption{The landscapes of the policy improvement objectives in ABR and TD3+BC under different regularization weight, fitted by neural networks over 2000 gradient steps. The size of the scatter is proportional to the possibility that the behavior policy selects the corresponding action. In TD3+BC, we observe that the scale of the regularization weight has a significant impact on the objective. A small weight is unable to mitigate the over-estimation of Q-function, while a large weight possibly instead leads to over-regularization on policy. In contrast, in ABR, due to the adaptive policy constraint weight, it effectively mitigates the over-estimation under a range of $\alpha$. Moreover, when the $\alpha$ is sufficiently small, the objective closely approximates the Q-function.} 
\label{multi-bandits}
\end{figure}

\textbf{Continuous Multi-arm Bandits}
We present a simple case to demonstrate the differences between ABR and previous methods that explicitly apply a policy constraint to the learned policy~\citep{fujimoto2021minimalist, wu2019behavior, kumar2019stabilizing, peng2019advantage, nair2020accelerating}. Specifically, consider a simple setting where an agent plays continuous one-dimension multi-arm bandits, where the action space is bounded by $[-1, 1]$. The agent has to learn its policy from experience obtained by a multimodal behavior policy. We compare ABR to TD3+BC~\citep{fujimoto2021minimalist}, which is one of the existing state-of-the-art methods. In TD3+BC, the policy improvement objective is Equation~(\ref{soft constraint}) with $f(s,a) = \mathbb{E}_{a'\sim \pi_{\beta}(\cdot|s)}\left[ (a - a')^2]\vert s\right]$. Figure~\ref{multi-bandits} shows the policy improvement objective in TD3+BC and ABR, respectively. 
As expected, ABR effectively mitigates the over-estimation out of the support of the behavior policy, which is relatively insensitive to the regularization coefficient. Meanwhile, ABR predicts the optimal Q-value near the true optimum of the collected actions, especially when the $\alpha$ is sufficiently small. In contrast, TD3+BC suffers from over-estimation when the $\alpha$ is small, while a large $\alpha$ instead results in over-regularization.

\textbf{In summary}, we highlight several significant advantages arising from the proposed adaptive policy constraint weight. (1) It provides an approximate identification of out-of-distribution actions, without explicitly estimating the behavior policy. Therefore, it enables ABR to handle datasets with multimodal or complex distribution. In our experiment, we empirically observe a substantial performance enhancement in complex datasets, compared to prior methods that enforce a constraint on policy (see Section~\ref{offline evaluation}). (2) The learned policy is updated following different patterns, dependent on the divergence between the learned policy and the behavior policy. Since the weight of the policy constraint term decreases monotonically as $\pi_{\beta}$ increases, the learned policy tends to clone the behavior policy when the divergence is large and improve the expected return when the divergence is slight. Thus, we effectively balance the trade-off described in Section~\ref{trade-off}.

\section{Experimental Evaluation}


In this section,  we aim to (1) introduce more implementation details of ABR and (2) evaluate ABR, compared with existing offline RL algorithms, on the D4RL~\citep{fu2020d4rl} benchmark tasks, which encompassed a range of domains and dataset compositions. In the hyperparameter study, we empirically demonstrate that ABR is easy to tune though several additional hyperparameters are incorporated. Furthermore, in our evaluation on offline tasks, ABR can outperform the existing state-of-the-art offline RL algorithms in most tasks and produce competitive performance in the rest of the tasks. We particularly focus on the comparison between ABR and prior methods that also aim to impose a constraint on policy. The experimental results strongly support that, compared to these methods, ABR is more adept at handling datasets with complex and multimodal distribution. 

\subsection{Implementations}
\textbf{Settings.} ABR can be easily implemented based on the standard actor-critic framework, with few lines modifications on codes. In practice, we build our algorithm on top of TD3 \citep{fujimoto2018addressing}, one of the existing state-of-the-art online RL algorithms. As discussed in Section~\ref{sec: ABR framework}, we prefer generalized energy distance to measure the discrepancy between $\pi_{\beta}$ and $\pi_{\theta}$, to avoid directly estimating $\pi_{\beta}$. Hence, we show the experiment results of ABR using generalized energy distance in the following evaluation. 
All experiments were performed on a server equipped with a single Tesla v-100 GPU and Intel(R) Core Xeon(R) CPU at 2.50GHz with 28 cores. See Algorithm~\ref{psuedocode} for a sketch of ABR\footnote{The implementation code of ABR will be released once the paper is accepted to publication.}. The performance on the D4RL benchmark is normalized from 0 to 100, corresponding to random and expert levels. 
\begin{algorithm}[t]
\caption{adaptive behavior regularization (generalized energy distance)}
\label{psuedocode}
  \SetAlgoLined
  Initialize Q-function $Q_{\phi}$ and a policy $\pi_{\theta}$.\
  
 \For{$k$ in $\{1, 2, \cdots, N\}$}
 {
    update $\phi$ by minimizing objective\
    
    $\mathbb{E}_{\substack{s \sim \mathcal{D} \\a\sim \pi_{\beta}(\cdot \vert s)}}
    \left[[Q_{\phi}(s,a)-\mathcal{B}^{\hat{\pi}^k}\hat{Q}^{k}(s,a)]^2 + 
    \alpha\mathbb{E}_{a'\sim \mathcal{U}}[Q_{\phi}(s,a')-\mathcal{B}^{\hat{\pi}^k}\hat{Q}^{k}(s,a) + \lambda(a-a')^2]^2\right]$.
    
    update $\theta$ by maximizing objective\\
    
    $\mathbb{E}_{\substack{s \sim \mathcal{D} \\a\sim \pi_{\theta}(\cdot \vert s)}} \left[\hat{Q}^{k+1}(s, a)\right]$.
 }
\end{algorithm}
To demonstrate the tuning details for ABR, we select two categories of tasks from the Gym domain of D4RL datasets (see a detailed introduction in Section~\ref{offline evaluation}), i.e., "-medium-expert" and "-medium-replay" tasks. We run each experiment for 1M gradient steps with four different random seeds and report the average performance.

\begin{figure}[t]
\centering
    \includegraphics[width=1.0\columnwidth]{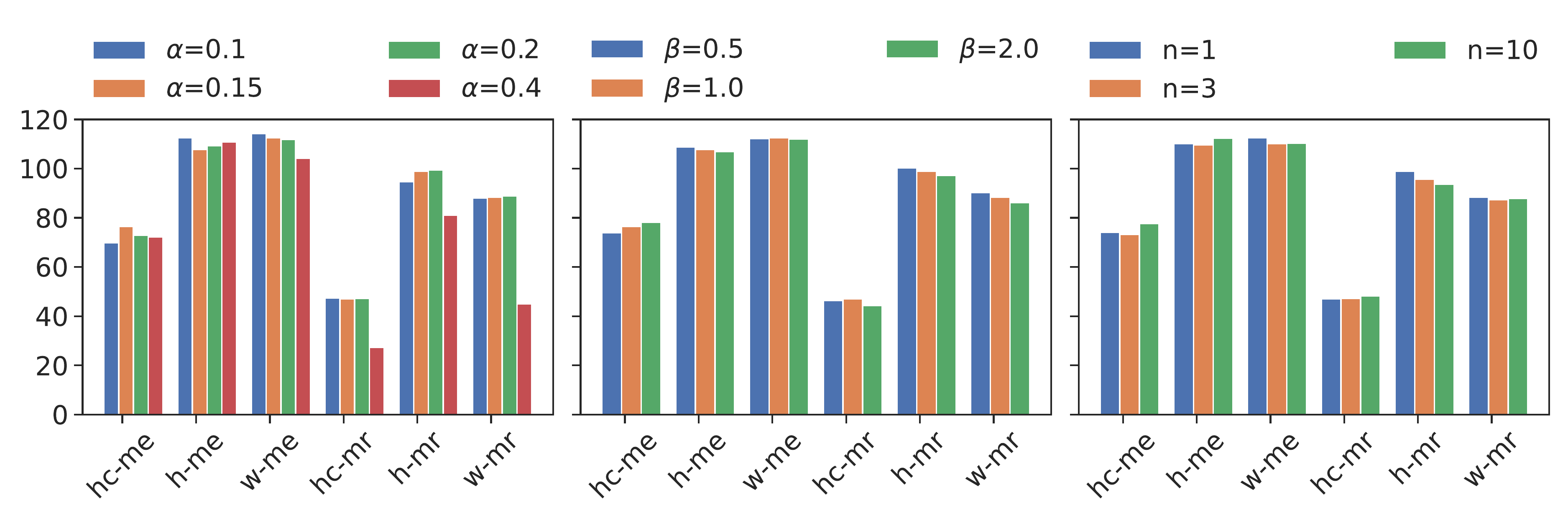}
\caption{Normalized score of ABR under different settings. The ticks of the horizontal axis represent different tasks, where hc=halfcheetah, h=hopper, w=walker2d, me=medium-expert, mr=medium-repaly. As expected, ABR performs similarly when $\alpha$ is sufficiently small (left). Besides, ABR is insensitive to $\beta$ (center). Hence, we can directly use the normalized distance in practice without further adjustment. Moreover, ABR performs well enough when only one action is sampled, which means only limited additional computational cost is required.\\}  
\label{fig hp}
\end{figure}

\textbf{Hyperparameter Studies.}  
First, we investigate the number of samples required to calculate the regularizer in ABR. The regularization term added to the Q-function estimation requires a uniform sampling over the entire action space. This seems computationally expensive since all sampled actions need to be incorporated in the backpropagation of the Q-function network. However, according to our experiment results, we empirically demonstrate that ABR requires only a few samples in each policy evaluation iteration. As shown in Figure~\ref{fig hp}, ABR with only \textit{one additional sample} can produce a similar performance compared to that with more samples. In the following evaluation on D4RL datasets, we default to sample only \textit{one} action in each policy evaluation iteration.

Besides, two additional hyperparameters are incorporated in ABR: the regularization coefficient $\alpha$ and the coefficient $\lambda$ of $f(s, a)$. As discussed in Section~\ref{sec: ABR framework}, we require a small $\alpha$ to reduce the bias in the Q-value of in-distribution actions. Thus we test ABR with $\alpha$ in the range of $[0.1, 0.2, 0.4]$. Meanwhile, we observe that the role of $\lambda$ is to balance the scale of Q-value and the $f(s, a)$ term. Hence, given that the action is bounded within $[a_{min}, a_{max}]$, we define the scalar $\lambda$ as: 
\begin{equation}
    \lambda=\beta \cdot \frac{(a_{max} - a_{min})^2}{\E_{\substack{s \sim \mathcal{D} \\a\sim \pi_{\beta}(\cdot \vert s)}}\lvert Q(s,a)\rvert}.
\end{equation}

A similar tuning trick is also utilized in the prior work~\citep{fujimoto2021minimalist}. We default to set $\alpha=0.15$ and $\beta=1.0$ in the subsequent experiment if no specific statement is given.
To demonstrate the effect of hyperparameters in ABR, we test different hyperparameter settings in MuJoCo locomotion tasks.
In Figure~\ref{fig hp}, we present the performance of ABR under different settings. 

We observe that when $\alpha<0.2$, ABR performs similarly in most tasks, while a larger one may result in underperforming in several tasks. Furthermore, the performance of ABR is insensitive to the choice of $\beta$. Thus we can simply tune it with a few attempts.

\subsection{Evaluation on Offline Benchmarks}
\label{offline evaluation}
\textbf{Baselines.} In D4RL benchmark~\citep{fu2020d4rl}, which comprises a range of continuous-control tasks and datasets, we compare the ABR in contrast to several prior algorithms, particularly those that enforce a constraint on policy. The baselines include prior methods that enforce a constraint on the policy: (1) BEAR~\citep{kumar2019stabilizing} that enforces an MMD distance constraint on the learned policy, (2) BRAC~\citep{wu2019behavior} that explicitly enforces a KL-divergence constraint, (3) AWR~\citep{peng2019advantage} that implicitly enforces a KL-Divergence constraint, (4) TD3+BC~\citep{fujimoto2021minimalist} that enforces an explicit energy distance constraint, and two existing state-of-the-art methods: (5) IQL~\citep{kostrikov2021offline} and (6) CQL~\citep{kumar2020conservative}. Also, we compare ABR to (7) behavior cloning (BC) and (8) online SAC~\citep{haarnoja2018soft}.

\textbf{Gym domains.}
 In Gym \citep{brockman2016openai} domains, the datasets are generated by first training policies online and collecting samples via the interaction with the Gym-Mujoco environment. According to the performance level of the trained policies, the datasets are labeled as ``expert" and ``medium" respectively. Moreover, ``medium-expert" datasets encompass the samples collected by multiple policies, and ``medium-replay" records the replay buffer~\citep{o2010play, lin1992reinforcement} of the medium level agent, both of whom construct a benchmark to evaluate the performance learned from heterogeneous and mixed experience.

Performance of ABR and baselines in the Gym domain are shown in Table~\ref{gym}. To ensure a fair comparison, all the normalized scores are based on the latest version of the datasets, namely `-v2'' datasets. We directly take the scores for IQL from the results reported by the author. As the scores reported by authors of BEAR, TD3+BC, and CQL are obtained by training in the earlier version, we re-train them respectively in the latest "-v2" datasets using author-provided opensource implementations\footnote{https://github.com/aviralkumar2907/CQL}\footnote{https://github.com/sfujim/TD3\_BC}\footnote{https://github.com/aviralkumar2907/BEAR}, and then report the results. This generally produces higher scores than that reported in original papers. In Table~\ref{gym}, we observe that ABR can surpass or match the performance of existing state-of-the-art methods in most tasks. Especially in the mixed datasets (“-medium-replay”, “-medium-expert”), ABR outperforms prior methods with a significant margin.

\begin{table}[t]\centering
\small
\caption{Normalized score of ABR and baselines on Gym domains from D4RL, averaged over four seeds. The scores that exceed 0.95x the best one are highlighted. ABR can match or exceed the performance of the best prior method in most tasks. Note that, on datasets with complex distribution ("medium-expert", "–medium-replay"), ABR significantly outperforms prior approaches that enforce a constraint on policy.}

\label{gym}
\renewcommand{\arraystretch}{1.1}
\newcolumntype{Y}{>{\centering\arraybackslash}X}
\begin{tabularx}{\linewidth}{l|YYYYY||Y}
\hline
Task Name                 & BC            & Bear                 & TD3+BC   & CQL   & IQL            & ABR   \\ \hline
halfcheetah-expert        &90.9        & 91.2                & \colorbox{table_color}{96.2}    & 72.4  & \colorbox{table_color}{94.7}           & 87.2           \\
hopper-expert             & \colorbox{table_color}{107.2} & 52.4                   & \colorbox{table_color}{108.9} & 101.1          & \colorbox{table_color}{110.0} & \colorbox{table_color}{107.4} \\
walker2d-expert           & \colorbox{table_color}{106.3}           & 99.9         & \colorbox{table_color}{110.2}  & \colorbox{table_color}{108.1} & \colorbox{table_color}{108.2} & \colorbox{table_color}{111.9} \\ \hline
halfcheetah-medium        & 42.8           & 39.7                  & 47.9          & 44.4 & 47.3          & \colorbox{table_color}{51.5}  \\
hopper-medium             & 54.5          & 36.9                 & 60.0          & \colorbox{table_color}{65.3} & \colorbox{table_color}{66.3}          & \colorbox{table_color}{67.8}           \\
walker2d-medium           & 73.1            & 3.1                  & \colorbox{table_color}{84.1} & 66.7 & 78.3          & \colorbox{table_color}{86.5}  \\ \hline
halfcheetah-medium-expert & 46.6          & 40.6                 & \colorbox{table_color}{89.1} & 63.3 & 86.7          & 76.5           \\
hopper-medium-expert      & 53.2           & 77.5                 & 96.7          & 98.0 & 91.5          & \colorbox{table_color}{107.8} \\
walker2d-medium-expert    & 94.7           & 12.8                  & \colorbox{table_color}{110.0} & \colorbox{table_color}{110.0} & \colorbox{table_color}{109.6} & \colorbox{table_color}{112.6} \\ \hline
halfcheetah-medium-replay & 36.1          & 37.6              & 44.6           & \colorbox{table_color}{45.4}  & 44.2           & \colorbox{table_color}{47.0}  \\
hopper-medium-replay      & 17.3          & 25.3                 & 50.2          & 90.6 & \colorbox{table_color}{94.7} & \colorbox{table_color}{98.9}  \\
walker2d-medium-replay    & 21.3          & 10.1                   & 80.7          & 79.6 & 73.9          & \colorbox{table_color}{88.3}  \\ \hline
average                   & 62.0           & 43.9                    & 80.2          & 78.7 & \colorbox{table_color}{82.8}          & \colorbox{table_color}{ 87.0}           \\
average of mixed task     & 44.9     & 34.0   & 78.6  & 81.2 & 83.4          & \colorbox{table_color}{88.5}           \\ \hline
\end{tabularx}

\label{gym}
\end{table}

\textbf{Adroit domains.}
The Adroit\citep{rajeswaran2017learning} domains encompass datasets with high-dimension representation and sparse reward, thus are much more complicated than Gym domains. The datasets labeled by``human" are collected by recording human behaviors. The ``clone" type datasets are generated by training a policy to imitate human demonstrations and mixing the samples collected by human behaviors and the cloned policy. We obtain the scores for the baselines from the official evaluation results of D4RL datasets~\citep{fu2020d4rl}. As shown in Table~\ref{adroit}, ABR significantly improves the scores compared with prior methods in several tasks. Moreover, compared with the methods that impose a constraint on the policy, ABR is the only method that effectively learns from the datasets.

\begin{table}[t]\centering
\small
\renewcommand{\arraystretch}{1.1}
\caption{Normalized score of ABR and baselines on adroit domains from D4RL, averaged over four seeds. The scores that exceed 0.95x the best one are highlighted. Note that ABR is the only policy constraint method that can effectively learn from the adroit tasks and significantly boost performance than simply behavior cloning.}
\label{adroit}

\newcolumntype{Y}{>{\centering\arraybackslash}X}
\begin{tabularx}{\linewidth}{l|ccccccc||Y}
\hline
Task Name       & SAC(online) & BC      & BEAR   & BRAC-v  & TD3+BC  & CQL           & IQL           & ABR \\
\hline
pen human       & 21.6        & 34.4    & -1     & 0.6    & -3.8    & 37.5 & 71.5 & \colorbox{table_color}{92.9}  \\
hammer human    & 0.2         & 1.5     & 0.3    & 0.2    & 1.5     & \colorbox{table_color}{4.4}           & 1.4           & 0.3            \\
door human      & -0.2        & 0.5     & -0.3   & -0.3   & 0       & \colorbox{table_color}{9.9}  & 4.3           & -0.1           \\
relocate human  & -0.2        & 0       & -0.3   & -0.3   & -0.3    & 0.2           & 0.1           & 0.1            \\
\hline
pen cloned      & 21.6        & 56.9    & 26.5   & -2.5   & -3.3    & 39.2 & 37.3 & \colorbox{table_color}{62.36} \\
hammer cloned   & 0.2         & 0.8     & 0.3    & 0.3    & 0.2     & \colorbox{table_color}{2.1}           & \colorbox{table_color}{2.1}           & 0.2            \\
door cloned     & -0.2        & -0.1    & -0.1   & -0.1   & -0.3    & 0.4           & \colorbox{table_color}{1.6}           & 0.2            \\
relocate cloned & -0.2        & -0.1    & -0.3   & -0.3   & 0.3     & -0.1          & -0.2          & 0.2            \\
\hline
average         & 5.35        & 11.73 & 3.13 & -0.30   & -0.71 & 11.7       & 14.76       & \colorbox{table_color}{19.52}    \\
\hline
\end{tabularx}
\\
\label{androit}
\end{table}

\section{Conclusions}
In this paper, we propose adaptive behavior regularization (ABR), a novel offline RL algorithm that achieves an adaptive balance between cloning and improving over the behavior policy. By simply adding a sample-based regularizer to the Bellman backup, we construct an adaptively regularized objective for the policy improvement, which implicitly estimates the probability density of the behavior policy.
Our method can be simply implemented based on the standard actor-critic framework. Empirically, we demonstrate that our method is easy to tune to meet an excellent performance. In our evaluation on the D4RL benchmark, our methods can match or surpass the performance of the best prior methods in most tasks. Furthermore, ABR can exceed prior policy constraint methods with a significant margin on datasets where the behavior policy is complex.

\bibliography{iclr2023_conference}
\bibliographystyle{iclr2023_conference}
\vfill
\newpage
\appendix

\section{Theoretical Proofs}
\label{theoretical proofs}
\textbf{Q-function Iteration.} 
For any $(s, a)$, by setting the deviation of the objective in Equation~\ref{Q-function iteration} to 0, we obtain an equation for $\hat{Q}_{\phi}^{k+1}(s,a)$:
\begin{equation}
\label{Q-function iter}
    \forall s \in \mathcal{D},a \in \mathcal{A} \quad  2\pi_{\beta}(a|s)\left[\hat{Q}_{\phi}^{k+1}(s,a)-\mathcal{B}^{\hat{\pi}^k}\hat{Q}_{\phi}^{k}(s,a)\right] + 
    2\alpha u \left[\hat{Q}_{\phi}^{k+1}(s,a)-\tilde{Q}(s,a)\right] = 0.
\end{equation}
Therefore, we obtain the expression of $\hat{Q}^{k+1}(s,a)$ by:
\begin{equation}
\begin{aligned}
\label{regularized Q-value 2}
\forall s \in \mathcal{D},a \in \mathcal{A} \quad
\hat{Q}_{\phi}^{k+1}(s,a)= &\frac{\pi_{\beta}(a|s)\mathcal{B}^{\hat{\pi}^k}\hat{Q}_{\phi}^{k}(s,a) + \alpha u \tilde{Q}(s,a)}{\pi_{\beta}(a|s)+\alpha u}\\
= & {\left(1 - \frac{\alpha u}{\pi_{\beta}(a|s)+\alpha u}\right)} \mathcal{B}^{\hat{\pi}^k}\hat{Q}_{\phi}^{k}(s,a) + \frac{\alpha u}{\pi_{\beta}(a|s)+\alpha u} \tilde{Q}(s,a).
\end{aligned}
\end{equation}

\textbf{Proof of Theorem~\ref{bounded bias}.}
Given that the reward $r$ is bounded by $r \in [-R_{max}, R_{max}]$, according to the definition of Q-value, we can always clip $\mathcal{B}^{\pi}Q(s,a)$ between $\left [-\frac{R_{max}}{1-\gamma}, \frac{R_{max}}{1-\gamma}\right ]$.
For any $s$, $a$, the bias introduced to standard Bellman backup satisfies:
\begin{equation}
\begin{aligned}
   & \left|\tilde{\mathcal{B}}^{\pi}Q(s,a) - \mathcal{B}^{\pi}Q(s,a) \right| \\
 = & \left|\frac{\alpha u}{\alpha u + \pi_{\beta}(a|s)}(\mathcal{B}^{\pi}Q(s,a)-\tilde{Q}(s,a)) \right|\\
\le & \frac{\alpha u}{\alpha u + \pi_{\beta}(a|s)} \left(\vert\mathcal{B}^{\pi}Q(s,a) \vert + \vert \tilde{Q}(s,a)\vert  \right)\\
 < & \frac{\alpha u}{\sigma}\left(\vert\mathcal{B}^{\pi}Q(s,a) \vert + \vert \tilde{Q}(s,a)\vert  \right)\\
 < & \frac{\alpha u}{\sigma} \left(\frac{R_{max}}{1-\gamma} + \delta\right),
\end{aligned}
\end{equation}
thus showing the bias can always be reduced by decreasing the scale of $\alpha$.
\section{Discussion of ABR}

\subsection{Reduction of Variance}
\label{variance}
We then discuss the variance of the evaluated Q-value in Equation~(\ref{Q-function iteration}). Since we add a regularizer to the policy evaluation objective, the variance of the evaluated Q-value is possibly increased.
We denote $\mathcal{B}^{\pi}\hat{Q}^k(s,a)$ as $Q_{\pi}^{k+1}(s,a)$ to simplify Equation~(\ref{Q-function iteration}). Accordingly, the right-hand side of Equation~(\ref{Q-function iteration}) can be rewritten as:
\begin{equation}
\label{Q-function iter}
\begin{aligned}
           & \mathbb{E}_{s\sim d^{\pi_{\beta}}, a\sim \pi_{\beta}} [Q(s,a)-Q_{\pi}^{k+1}(s,a)]^2 + \alpha\mathbb{E}_{s\sim d^{\pi_{\beta}},a\sim \mathcal{U}}[Q(s,a)-\tilde{Q}(s,a)]^2\\
           & = \mathop{\int}_{s}d^{\pi_{\beta}}(s)\mathop{\int}_{a}\pi_{\beta}(a|s)[Q(s,a)-Q_{\pi}^{k+1}(s,a)]^2 + \alpha u [Q(s,a)-\tilde{Q}(s,a)]^2  dads\\
           & = 2\mathop{\int}_{s}d^{\pi_{\beta}}(s)\mathop{\int}_{a}\frac{\pi_{\beta}(a|s)+\alpha u}{2}
           \left( \frac{\pi_{\beta}(a|s)}{\pi_{\beta}(a|s)+\alpha u}[Q(s,a)-Q_{\pi}^{k+1}(s,a)]^2 + \frac{\alpha u}{\pi_{\beta}(a|s) + \alpha u} [Q(s,a)-\tilde{Q}(s,a)]^2 \right) da ds.
\end{aligned}
\end{equation}
For each $(s, a)$, we define a variable $y(s,a)$ following a distribution whose probability density function $p$ over possible outcomes $y(s,a)$ is as follows:
\begin{equation}
\label{y(s,a)}
    \begin{aligned}
    \forall s, a \quad p(y(s,a)) = 
    \begin{cases}
      \frac{\pi_{\beta}(a|s)}{\pi_{\beta}(a|s)+\alpha u}, & for \quad y=Q_{\pi}^{k+1}(s,a)\\ 
      \frac{\alpha u}{\pi_{\beta}(a|s) + \alpha u} , & for\quad y = \tilde{Q}(s,a)
    \end{cases}
    \end{aligned}
\end{equation}
Since $\mathop{\int}_{a}\frac{\pi_{\beta}(a|s)+\alpha u}{2} da=1$, it can be regarded as we sample $a$ for given $s$ following a distribution whose possibility density function is $p(s,a)=\frac{\pi_{\beta}(a|s)+\alpha u}{2}$. The training signal of $Q(s,a)$ can be provided either by minimizing the $[Q(s,a) - Q_{\pi}^{k+1}(s,a)]^2$ or minimizing $[Q(s,a) - \tilde{Q}(s,a)]^2$. In each iteration, we obtain the evaluated Q-value for any $(s, a)$ by fitting the expectation of $y(s,a)$ following this distribution. Intuitively, if we greedily obtain the optimum for Equation~\ref{Q-function iter}, it resembles a supervised learning with mixed labels, thus the regularizer possibly increase the variance for fitting the Q-function. According to Equation~\ref{y(s,a)}, the variance of $y(s,a)$:
\begin{equation}
\begin{aligned}
    \forall s, a \quad Var(y(s,a)) = &\frac{\alpha u\pi_{\beta}(a|s)}{\pi_{\beta}(a|s)+\alpha u}[Q_{\pi}^{k+1}(s,a)-\tilde{Q}(s,a)]^2\\
                                   = &  \frac{\alpha u\pi_{\beta}(a|s)}{\pi_{\beta}(a|s)+\alpha u}[Q_{\pi}^{k+1}(s,a)-c(s) + f(s,a)]^2.
\end{aligned}
\end{equation}
We can reduce this variance by setting an appropriate $c(s)$. For any $s$, the expectation of this variance is given as:
\begin{equation}
\begin{aligned}
    \forall s \quad &\E_{a\sim \frac{\pi_{\beta}(a|s)+\alpha u}{2}} \left[Var(y(s,a))\middle|s\right]\\
    = &  \mathop{\int}_{a}\frac{\pi_{\beta}(a|s)+\alpha u}{2} \cdot \frac{\alpha u\pi_{\beta}(a|s)}{\pi_{\beta}(a|s)+\alpha u}[Q_{\pi}^{k+1}(s,a)-c(s) + f(s,a)]^2  da\\
    = & \frac{\alpha u}{2} \mathop{\int}_{a} \pi_{\beta} [Q_{\pi}^{k+1}(s,a)-c(s) + f(s,a)]^2  da\\
    = & \frac{\alpha u}{2} \mathop{\int}_{a} \pi_{\beta} [(Q_{\pi}^{k+1}(s,a)-c(s)) ^ 2 + f(s,a)^2 + 2f(s,a)(Q_{\pi}^{k+1}(s,a)-c(s))]  da \\
    \le & \frac{\alpha u}{2} \mathop{\int}_{a} \pi_{\beta} [(Q_{\pi}^{k+1}(s,a)-c(s)) ^ 2 + f(s,a)^2 ] 
\end{aligned}
\end{equation}



     


\end{document}